\documentclass{article}

\usepackage{arxiv}

\usepackage[utf8]{inputenc} % allow utf-8 input
\usepackage[T1]{fontenc}    % use 8-bit T1 fonts
\usepackage{hyperref}       % hyperlinks
\usepackage{url}            % simple URL typesetting
\usepackage{booktabs}       % professional-quality tables
\usepackage{amsfonts}       % blackboard math symbols
\usepackage{nicefrac}       % compact symbols for 1/2, etc.
\usepackage{microtype}      % microtypography
\usepackage{graphicx}
\usepackage{natbib}
\usepackage{doi}
\usepackage{algorithm}      % algorithm package added
\usepackage{algpseudocode}  % pseudocode package for algorithms
\usepackage{amsmath}
\usepackage{amssymb}
\usepackage{mathtools}
\usepackage{amsthm}
\usepackage{subfigure}      % Added to support subfigures

\usepackage[utf8]{inputenc} % allow utf-8 input
\usepackage[T1]{fontenc}    % use 8-bit T1 fonts
\usepackage{hyperref}       % hyperlinks
\usepackage{url}            % simple URL typesetting
\usepackage{booktabs}       % professional-quality tables
\usepackage{amsfonts}       % blackboard math symbols
\usepackage{nicefrac}       % compact symbols for 1/2, etc.
\usepackage{microtype}      % microtypography
\usepackage{lipsum}		% Can be removed after putting your text content
\usepackage{graphicx}
\usepackage{natbib}
\usepackage{doi}

\theoremstyle{plain}

\theoremstyle{definition}

\theoremstyle{remark}

\title{COMPRER: A Multimodal Multi-Objective Pretraining Framework for Enhanced Medical Image Representation}

\author{
    Guy Lutsker$^1$ \\
    \texttt{guy.lutsker@weizmann.ac.il} \\
    \And
    Hagai Rossman$^2$ \\
    \texttt{hagai@pheno.ai} \\
    \And
    Nastya Godiva$^1$ \\
    \texttt{nastya.godiva@weizmann.ac.il} \\
    \And
    Eran Segal$^{1,3}$ \\
    \texttt{eran.segal@weizmann.ac.il} \\
}

\begin{document}
\maketitle

\footnotetext[1]{Department of Math and Computer Science, Weizmann Institute of Science, Rehovot, Israel}
\footnotetext[2]{Pheno.AI, Tel Aviv, Israel}
\footnotetext[3]{Department of Machine Learning, Mohamed Bin Zayed University of Artificial Intelligence, Abu Dhabi, UAE}

% keywords can be removed
%\keywords{Machine Learning \and Medical Imaging \and %Multimodal Learning}

\begin{abstract}
%(COntrastive Multi-objective PREtraining for multi-modal Representation)%

Substantial advances in multi-modal Artificial Intelligence (AI) facilitate the combination of diverse medical modalities to achieve holistic health assessments. We present COMPRER , a novel multi-modal, multi-objective pretraining framework which enhances medical-image representation, diagnostic inferences, and prognosis of diseases. COMPRER employs a multi-objective training framework, where each objective introduces distinct knowledge to the model. This includes a multimodal loss that consolidates information across different imaging modalities; A temporal loss that imparts the ability to discern patterns over time; Medical-measure prediction adds appropriate medical insights; Lastly, reconstruction loss ensures the integrity of image structure within the latent space. Despite the concern that multiple objectives could weaken task performance, our findings show that this combination actually boosts outcomes on certain tasks. Here, we apply this framework to both fundus images and carotid ultrasound, and validate our downstream tasks capabilities by predicting both current and future cardiovascular conditions. COMPRER achieved higher Area Under the Curve (AUC) scores in evaluating medical conditions compared to existing models on held-out data. On the Out-of-distribution (OOD) UK-Biobank dataset COMPRER maintains favorable performance over well-established models with more parameters, even though these models were trained on $75\times$ more data than COMPRER. In addition, to better assess our model's performance in contrastive learning, we introduce a novel evaluation metric, providing deeper understanding of the effectiveness of the latent space pairing.

\end{abstract}

\section{Background}

\begin{figure*}[!h]
\centering
\includegraphics[ width=1\textwidth]{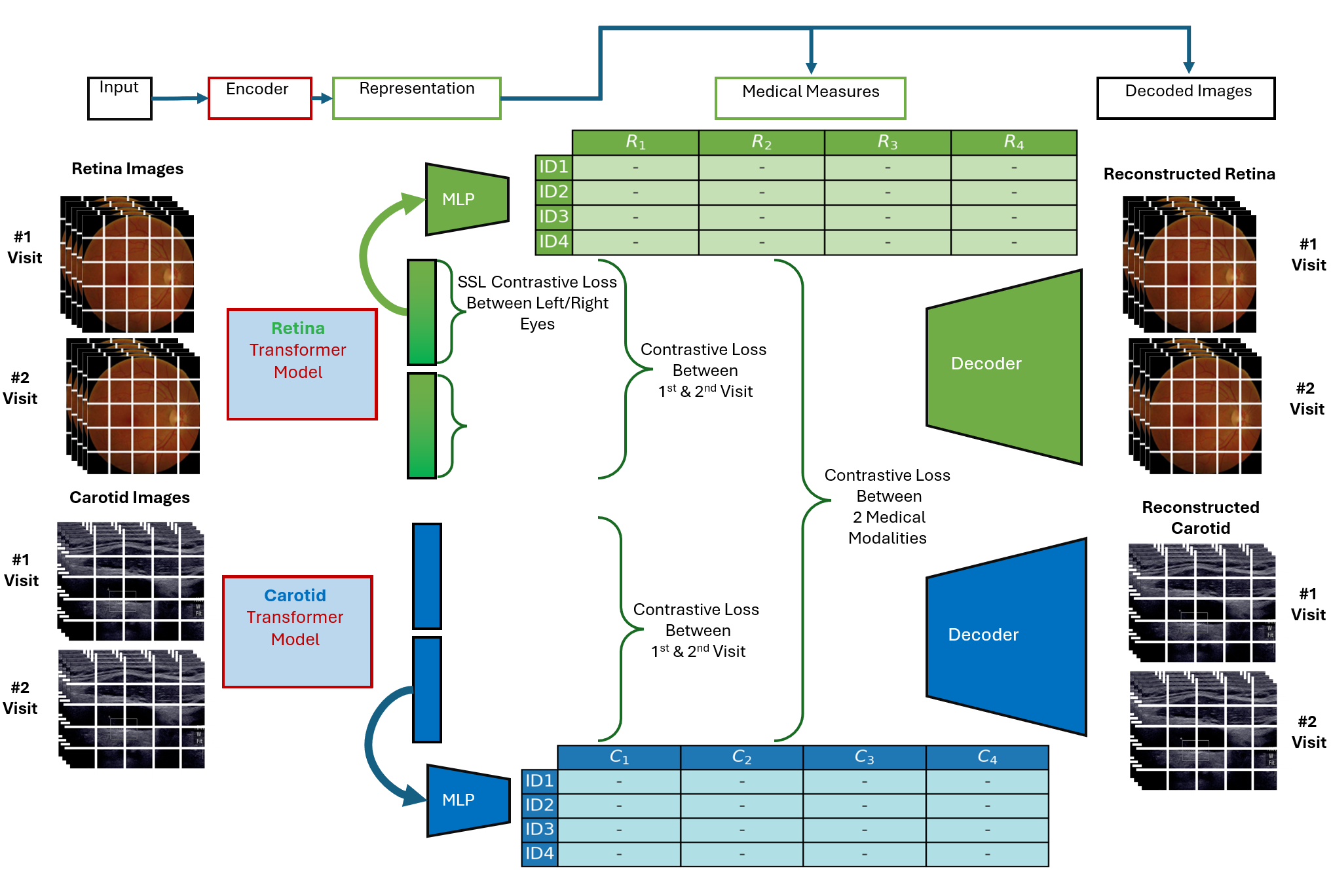}
\caption{Schematic representation of COMPRER, our Contrastive Multi-objective Pretraining approach for Multi-modal Representation. We utilize ViT-Base encoders equipped with DINOV2 pre-trained weights for processing each imaging modality, accompanied by a linear projection head. Our method is defined by a combination of multiple loss objectives: (1) an ID-centric multi-modal contrastive loss that bridges features between fundus images and carotid ultrasound images; (2) a patient visit-based contrastive loss that discerns temporal discrepancies across repeat visits for each patient; (3) a contrastive scheme for the bilateral fundus images to ensure coupling of right and left eye data per patient; (4) a decoding objective to restore original images from condensed latent representations; and (5) a predictive mechanism to estimate general medical measures directly from modality-specific embeddings.}
\label{fig.1}
\end{figure*}
The evolution of AI within healthcare is promoting an era of precision medicine, marked by enhanced diagnostic accuracy, improved prognostic evaluations, and personalized treatment strategies \citep{Bajwa2021, Esteva2019}, where deep learning is increasingly central to medical imaging analysis \citep{Huang2023}. Technologies such as fundus imaging and carotid ultrasound are pivotal in cardiovascular health assessments, granting insights into micro and macrovascular structures and pathologies \citep{Poplin2018, Spence2006}. Fundus imaging, a non-invasive procedure, reveals the retinal microvasculature and is used to detect early manifestations of diseases like diabetes and hypertension \citep{Dai2021, Yan2019}. Such microvascular changes are significant indicators of systemic conditions, enabling broader health monitoring. Carotid ultrasound complements  fundus imaging by providing a structural assessment of the carotid arteries, crucial for identifying risks of stroke and atherosclerosis through blood flow dynamics and plaque visualization \citep{Yu2021, Siontis2021}. The integration of both the fundus imaging, and carotid ultrasound modalities, offers a comprehensive representation of cardiovascular health, capitalizing on their individual strengths to assess conditions at various scales of the vascular system. However, the full potential of AI in medical imaging is challenged by the limited availability of large, annotated datasets necessary for traditional supervised learning \citep{Zhou2023}. This dataset scarcity is addressed by initiatives such as the Human Phenotype Project (HPP), which embraces a multi-modal deep-phenotyping approach, capturing a vast range of data modalities, from high-resolution images to comprehensive clinico-pathological records \citep{Shilo2021}. Such datasets are ideal for investigating and improving AI models that surpass existing boundaries in medical diagnostic capabilities. \citep{Moor2023}. To overcome these limitations 
self-supervised learning (SSL) has become a key tool in this field. \citep{Huang2023}. SSL circumvents the need for extensive labeled datasets by utilizing the data itself to derive informative features through the resolution of proxy tasks. It enables the extraction of significant patterns intrinsic to the data, fostering a model's ability to generalize robustly to unseen data \citep{Grill2020, Chen2020}. SSL with multi-modal data are particularly potent, with each modality enhancing the model's capabilities and utility across various health signals \citep{Radhakrishnan2023}.

\section{Introduction}
In this paper, we present COMPRER (COntrastive Multi-objective PREtraining for multi-modal Representation), a multi-modal, multi-objective pretraining framework that can be used in numerous downstream tasks and applications through the analysis of fundus imaging and carotid ultrasound. These include: diagnosing current patient diseases, predicting clinically significant medical features, and prognosing the probability of developing a medical condition in the future. As seen in Figure 1, COMPRER using a multi-modal approach, where it integrates distinct but complementary data sources— fundus imaging and carotid ultrasound imaging. Each modality offers a unique glimpse into the cardiovascular health of patients, capturing a diverse array of medical measures that, when combined, provide a comprehensive assessment framework. Our framework leverages ViTs \citep{Dosovitskiy2020}, specifically the DINOV2-Base pre-trained model \citep{Oquab2023}, as our architectural backbone, augmented by a multi-objective learning strategy that incorporates reconstruction via an image decoder, predictive heads, and contrastive learning losses.

\subsection{COMPRER Training Objectives}
In this section, we describe the training objectives utilized for our model. Our approach employs paired batches of fundus images and carotid ultrasound images to learn a joint embedding space, inspired by the CLIP training paradigm \citep{Radford2021}. This multimodal training maximizes the similarity of embeddings from matching image pairs and minimizes it for non-matching pairs within a batch.

The contrastive loss function specific to a set of two embeddings types \(u\) and \(v\) is defined as:
\[
\mathcal{L}_{\text{contr}}(u, v) = -\frac{1}{N}\sum_{i=1}^{N}\log\left(\frac{\exp\left(\frac{\text{sim}(u_i, v_i)}{\tau}\right)}{\sum_{j=1}^{N}\exp\left(\frac{\text{sim}(u_i, v_j)}{\tau}\right)}\right)
\]
where \(\text{sim}(u,v) \equiv \frac{u^{\top}v}{\|u\|\|v\|}\) and \(N\) is the batch size. The contrastive loss is as in CLIP: \\\(\mathcal{L}_{\text{contr\_CLIP}}(u,v) =  \frac{1}{2}\left(\mathcal{L}_{\text{contr}}(u, v) + \mathcal{L}_{\text{contr}}(v, u) \right)\)\\Several contrastive losses based on the relationship between the embeddings compared:
For fundus (\(f\)) and carotid (\(c\)) image embeddings: 
\(\mathcal{L}_{\text{contr\_fc}} = \mathcal{L}_{\text{contr\_CLIP}}(f, c)\).

For fundus images across different visits \(t\) and \(t'\): 
\(\mathcal{L}_{\text{contr\_fv}} = \mathcal{L}_{\text{contr\_CLIP}}(f^{t}, f^{t'})\) \\
For carotid images over time \(t\) and \(t'\): 
\(\mathcal{L}_{\text{contr\_cv}} = \mathcal{L}_{\text{contr\_CLIP}}(c^{t}, c^{t'})\) \\
For right (\(R\)) and left (\(L\)) eye fundus images: 
\(\mathcal{L}_{\text{contr\_eye}} = \mathcal{L}_{\text{contr\_CLIP}}(f^{R}, f^{L})\)

Predictive accuracy for fundus and carotid embeddings is assessed with Mean Squared Error (MSE) losses:
\[
\mathcal{L}_{\text{pred}}(m, \hat{m}) = \frac{1}{N}\sum_{i=1}^{N}(m_i-\hat{m}_i)^{2}
\]
Yielding \(\mathcal{L}_{\text{pred\_r}}\) for fundus with \(N=N_r\), and \(\mathcal{L}_{\text{pred\_c}}\) for carotid with \(N=N_c\), with their respective medical measurements predictions. Finally, the total loss \(\mathcal{L}\) combines individual contrastive and predictive components:
\[
\mathcal{L} = \mathcal{L}_{\text{contr\_fc}} + \mathcal{L}_{\text{contr\_fv}} + \mathcal{L}_{\text{contr\_cv}} + \mathcal{L}_{\text{contr\_eye}} + \mathcal{L}_{\text{pred\_r}} + \mathcal{L}_{\text{pred\_c}},
\]

\subsection{Model Summary and Key Contributions}
We evaluate COMPRER through validation across individual learning objectives and demonstrate its capabilities not only in extracting meaningful representations but also in projecting these representations into actionable clinical insights. Notably, we show that our multi-objective framework results in enhanced diagnostic and prognostic accuracy. Moreover, the model’s ability to outperform not only a baseline pretrained DINOV2 but also dedicated models with substantial advantages in terms of parameters and data scale demonstrates the efficacy of our approach. In conclusion, our contributions are:
\begin{enumerate}
  \item We introduce COMPRER, a novel deep learning framework that leverages multi-modal, multi-objective pretraining to forecast and predict the development of future diseases from medical imaging data.
  \item We provide evidence for the efficacy of our modeling approach through an internal validation scheme, showing that our embeddings are capable of predicting medical measures with high R\textsuperscript{2} scores.
  \item We introduce a novel, understandable metric for assessing the performance of contrastive learning, offering an approach to measure the quality of embeddings in identifying correct image pairs across different modalities.
  \item We substantiate the translational value of our model through its application in predicting cardiovascular health conditions,  in both our cohort as well as in external cohorts.
\end{enumerate}

% Continue adding other paragraphs and sections as needed.
\section{Related Work}

The combination of AI with healthcare represents a significant shift toward redefining clinical methodology and patient care. At the heart of this transformation, deep learning, particularly through convolutional neural networks (CNNs), has played a pivotal role in enhancing medical diagnostics. CNNs have demonstrated their efficacy in detection and classification tasks across a range of medical imaging modalities, including dermatology \citep{Kwasigroch2020}, radiology \citep{Tiu2022, Jaiswal2019}, and neuroradiology \citep{Pereira2016}. The profound pattern recognition capabilities of CNNs have thus become instrumental in medical image interpretation. Recently, the success of SSL methods, such as SimCLR \citep{Chen2020} and BYOL \citep{Grill2020}, has redirected the focus from supervised learning reliant on extensive labeled datasets to SSL in extracting features from unlabeled data \citep{Huang2023}. SSL's resilience to dataset imbalances \citep{LiuH2021} particularly proves its adaptability in medical contexts, making it a cornerstone for foundation models, designed for broad application across multiple tasks \citep{Bommasani2021, Moor2023}. Zhou et al. \citep{Zhou2023} exemplified this adaptability in a self-supervised masking strategy applied to a vast array of unlabeled fundus images, yielding a foundation model with an ability to perform disease detection across multiple scenarios. The breakthrough with OpenAI's Contrastive Language-Image Pretraining (CLIP) system has provided a novel perspective on utilizing versatile architectures, specifically transformers \citep{Vaswani2017}, for a wide range of modalities \citep{Radford2021, Ramesh2022, Ramesh2021, Brown2020}. CLIP's revolutionary approach to interpreting images through a natural language lens has revealed the potential of transformer architectures to tokenize and process multimodal data efficiently. Our COMPRER framework adopts a similar stance, leveraging the transformer architecture for both image encoders in a contrastive mechanism to align information across medical imaging types. The methodology rooted in CLIP's cross-modal learning inspired both the multi-visit and multi-modal contrastive losses of COMPRER, allowing it to not only decode spatial characteristics but also trace temporal patterns indicative of disease progression. In the era of multimodal data, significant strides have been made in cross-modal representations \citep{Radhakrishnan2023} and feature extraction \citep{Holmberg2020}, revealing the intersecting pathways of SSL and multimodal methodologies. Complementing this trend are ViTs, which have remodeled the AI landscape with their extraordinary image processing capabilities \citep{Dosovitskiy2020} and interpretability \citep{Chefer2020}. The advent of multi-task learning frameworks further amplifies these models' ability to assimilate diverse data and objectives, showcasing their robustness across a spectrum of tasks relevant to cardiovascular health analytics \citep{Ruder2017, Crawshaw2020}.

\section{Methodology}
In our study, we present a multimodal, multi-objective deep learning architecture designed to create a versatile pretrained model suitable for a wide range of health-related tasks. This architecture can generate adaptable embeddings for predicting a multitude of medical features or be fine-tuned for diverse medical applications. The model achieves this through the analysis of both fundus images and carotid ultrasounds. This part delves into the intricacies of our approach, which capitalizes on the robust capabilities of ViTs. Initially, we assembled a dataset encompassing approximately 11.5K participants' fundus and carotid ultrasound images, of which 1.5K have returned for a follow-up visit after two years. The dataset is divided into training (80\% of data), validation (validation is 20\% of the training set), and test sets (20\% of the data), with the latter consisting solely of new participants arriving after the start of this research to ensure the integrity of our evaluation. Our preprocessing protocols ensure high-quality, artifact-free images using AutoMorph \citep{zhou2022automorph} for fundus images and custom preprocessing to isolate relevant regions in carotid ultrasounds. Both image types are standardized to a resolution of 280x280 pixels, facilitating uniform processing where even grayscale ultrasound images are converted to three-channel format to align with the fundus images. The structural backbone of our model derives from the pretrained DINOV2-Base ViT, a vision transformer by Meta that has shown great performance in image representation tasks, as well as multiple vision downstream tasks \citep{Caron2021, Oquab2023}. As DINOV2 was trained on millions of images, we can capitalize on its extensive pretrained ability to represent image datasets and we can fine-tune it to our unique medical imaging context for enhanced efficiency. To complement the ViT, a linear projection head condenses high-dimensional embeddings to a more manageable state, serving a dual purpose: reducing computational demands and assisting with stability and shown to be essential by \citet{Balestriero2023}. A transposed convolution neural network, comprising of transposed convolutional layers with gaussian error linear unit (GELU) activations, reconstructs the original images from latent embeddings, introducing a regularization effect that underpins the self-supervised learning within our framework. Lastly, a small 2-layer Multi-Layer Perceptron (MLP) is added to predict medical measures from the latent space embeddings. To train the model we use multiple different optimization objectives simultaneously. These include the multimodal and multi-visit contrastive losses used as shown in CLIP by OpenAI \citep{Radford2021}, which fortify the model's time-awareness and cross-modality inference capabilities. A classic mean squared error (MSE) decoder loss ensures fidelity in image reconstruction, and another MSE loss is applied when predicting medical measures from the embeddings. We trained the model over four days, distributed across 8 NVIDIA A40 GPUs. We trained with an AdamW optimizer, with learning of $3\times10^{-4}$ and weight decay of $0.5$ with a StepLR scheduler, and a batch size of 9 per GPU. We found that the model does not overfit the train set in this time period, but due to limited resources we chose to run for only 4 days (even though the model might not have saturated the training set during this period). Addressing missing data, we introduce four parallel data loaders that guarantee optimal usage of the available data by including every sample where possible in the training, even when dealing with missing modalities or visit data. During training, we employ the validation set to discern the model's evolving accuracy. To evaluate performance on the validation set, we used different metrics for the different losses we employed. For the medical measures task, we relied on the R\textsuperscript{2} to gauge the medical measure predictions from latent embeddings. For the decoding task, we relied on the decreasing MSE loss, as well as human evaluation of the resulting reconstructions. To evaluate the contrastive learning performance, we have crafted a novel metric, assessing the proximity of paired image embeddings and optionally adjusting for random chance, thus providing an intuitive measure of the model's learning. Essentially, by viewing the contrastive task as a classification task, we can view this metric as top-K accuracy. 

\algtext*{EndIf}% Remove "end if" text
\algtext*{EndFor}% Remove "end for" text
\algtext*{EndProcedure}% Remove "end procedure" text
\algtext*{EndWhile}% Remove "end while" text

% Redefine the For and While commands with a colon
\renewcommand{\algorithmicfor}{\textbf{for}}
\renewcommand{\algorithmicdo}{:}
\renewcommand{\algorithmicwhile}{\textbf{while}}

\vspace{-3mm}
\begin{algorithm}[H]
\caption{Top-K Metrics for Contrastive Learning}\label{alg:contrastive_metric}
\fontsize{10pt}{9pt}\selectfont % This sets the font size to small. Replace with your desired size.
\begin{algorithmic}%[1]

\Procedure{TopK}{$sim\_mat$, $k$} 
  \State $correct \gets 0$
  \For{$i = 0$ \textbf{to} $|sim\_mat|$}
    \If{index $i$ in top-$k$ similar items}
      \State $correct \gets correct + 1$
    \EndIf
  \EndFor
  \State \textbf{return} $correct \text{ } /\text{ }  |sim\_mat|$
\EndProcedure
\State  

\State $kVals \gets [5, 25, 100, \ldots]$
\State $emb_i, emb_j \gets \text{embeddings}$
\State $norm_i \gets \text{l2\_normalize}(emb_i)$
\State $norm_j \gets \text{l2\_normalize}(emb_j)$
\State $cosSimMat \gets norm_i \cdot norm_j^\top$

\For{$k \in kVals$}
  \State $rand\_base \gets  k \text{ } / \text{ } |cosSimMat|$
  \State $metricScore \gets \Call{TopK}{cosSimMat, k}$
  \State $angleTopK \gets metricScore$
  \State $multAngleK \gets metricScore / randBase$
\EndFor

\end{algorithmic}
\end{algorithm}

In contrastive learning, the central goal is to learn representations such that similar or "paired" samples are brought closer together in the embedding space, while dissimilar samples are pushed apart. As we deal with batches of N samples, we inherently face an N-way classification problem during training. Achieving perfect performance is often challenging, and a binary assessment of model proficiency via top-1 prediction accuracy may not sufficiently capture the nuances in the embeddings the model has learned. In practice, it may appear that the model is underperforming when, in fact, it has developed a representation where correct matches are amongst the nearest neighbors, not necessarily the immediate first. By introducing a Top-K metric (Algorithm 1) specifically tailored for contrastive learning, we extend the single-label evaluation to a multi-neighbor perspective, which is analogous to considering a set of K nearest neighbors in the embedding space. Selecting different values of K enables us to explore the depth of the model's understanding of data relationships. Lower values of K can indicate fine-grained discriminatory power, while larger values suggest a broader comprehension of sample similarity. Furthermore, by adjusting for random chance in our metric — by dividing the raw Top-K score by the expected score under random matching — we gain insight into how much more effectively our model is at reconciling these pairs compared to a trivial random embedding model. Notably, this also allows us to use our hardware efficiently, as to avoid evaluating our model directly using the downstream tasks, we can evaluate the contrastive task performance at pretraining time.

\begin{figure}[!ht]
\begin{center}
\includegraphics[scale=0.32]{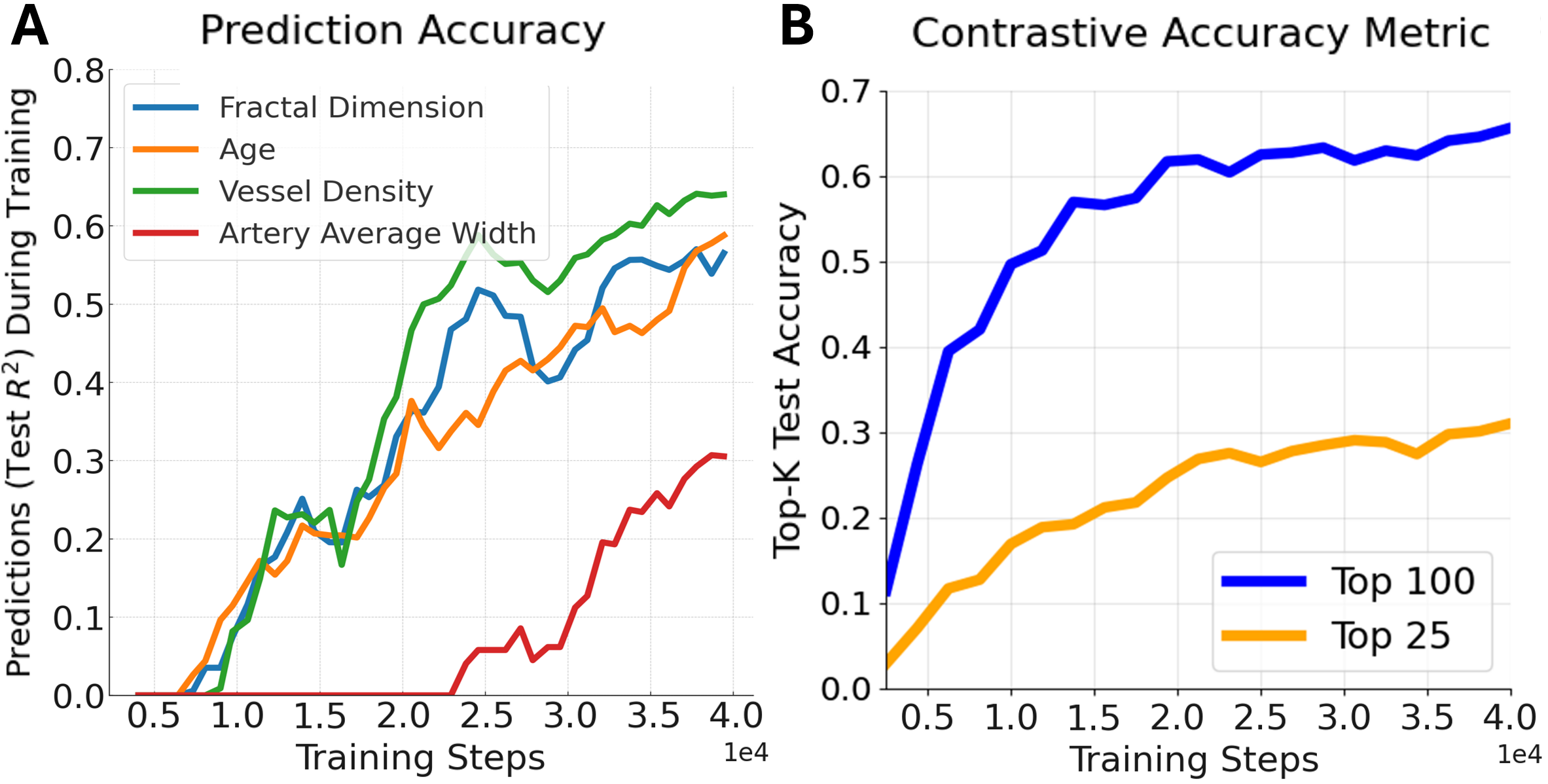} 
\caption{Interval Validation, \textbf{2.A} (left panel). Prediction accuracy of various medical measures over iterative steps. This figure illustrates the test R\textsuperscript{2} values achieved during the training of predictive models for four different medical measures: fractal dimension, age, vessel density, and artery average width. The x-axis is the training steps iterations (in tens of thousands), while the y-axis indicates the R\textsuperscript{2} value observed on held-out test data.  \textbf{2.B} (right panel). Evolution of Top-K Test Accuracy in Contrastive Learning for Multimodal Image Matching. This figure depicts the test accuracy of a contrastive learning model for two different values of k: 25 and 100, as labeled Top-25 (orange line) and Top-100 (blue line), respectively. The x-axis is, again, training steps iterations, and the y-axis denotes the Top-K Test Accuracy. The Top-100 accuracy increases as training goes on, reaching top performance of 0.65. Similarly, the Top-25 accuracy ends around 0.35. }
\label{fig.2}
\end{center}
\end{figure} 

\section{Results}
\subsection{Internal Validation}
Our exploration of COMPRER's results began with internal validation metrics. Interval validation metrics are composed of metrics that validate the pretraining phase of our model with multiple objectives. For each one of these objectives, we report their score on a held-out test set.
\subsubsection{Medical Measures Prediction}
To evaluate the generalization capabilities of the medical measures prediction head of COMPRER, we engaged in predicting measures from the test set that have direct clinical applicability. Among the predicted medical measurements were age, fundus image fractal dimension, vessel density, and artery average width. Predicting age from fundus images is particularly intriguing, as it suggests a correlation between ocular characteristics and biological aging, which can have various medical implications \citep{ahadi2023longitudinal}. The Fundus Image Fractal Dimension is a measure of the complexity and branching patterns of the retinal vasculature, indicative of overall vascular health\citep{dinesen2021retinal,macgillivray2007fractal} . Vessel Density refers to the proportion of the retina occupied by blood vessels, a crucial factor in assessing retinal and systemic circulatory health. Artery Average Width provides insights into vascular caliber, important for understanding cardiovascular risks. The ability to predict these measures from fundus images is noteworthy, indicating that our model retains spatial understanding of the images despite the multiple objectives enforced on this image representation. As detailed in Figure 2, the generalization performance, quantified by R\textsuperscript{2} score — showed differing degrees of success across the medical measures. An R\textsuperscript{2} score of approximately 0.6 was observed for most medical measures, indicating a meaningful predictive relationship between the learned representations and the clinical measures. However, the prediction for Artery Average Width presented a lower R\textsuperscript{2} score of around 0.3, signifying a less robust prediction capability, or a harder prediction task.
\subsubsection{Evaluation of Contrastive Learning}
A critical component of the COMPRER framework is the multimodal contrastive loss, which plays a pivotal role in aligning features across distinct imaging modalities—namely, fundus and carotid ultrasound images. In fact, in the whole multiple objective framework, this loss is the only connection between the two distinct image encoders. To measure the effectiveness of contrastive learning, we devised a scaleless, interpretable metric that provides a concrete understanding of model performance. While the loss term itself provides an indication of model learning during training, it lacks direct translatability to practical outcomes. In Figure 3.a, we show the Top-K test accuracy of $k \in [5, 25,100]$, and in Figure 3.b we show the same plot for the multiplicative metric, which highlights the model getting better than random performance. The plots show the metric as calculated from algorithm 1. The model exhibited non-trivial, non-random results in the multimodal matching task, which is notable considering the inherent challenge of this problem - one that even skilled clinicians do not typically address. As part of our experimental setup, we developed an ablation model, the Multi-Modal Contrastive Learning (MMCL) model, which was trained on the same data as COMPRER. Unlike COMPRER, MMCL was trained using a single objective with pretraining - focusing exclusively on multi-modal contrastive learning. This approach allowed us to evaluate the impact of the other objectives on our main model's performance. Interestingly, the COMPRER model outperforms the Multi-Modal Contrastive Learning (MMCL) model, which has trained on the same data, with only the multi-modal contrastive loss, in multimodal matching accuracy, emphasizing the advantage of a multiple objective training strategy. 
\begin{figure}[!ht]
\begin{center}
\includegraphics[scale=0.32]{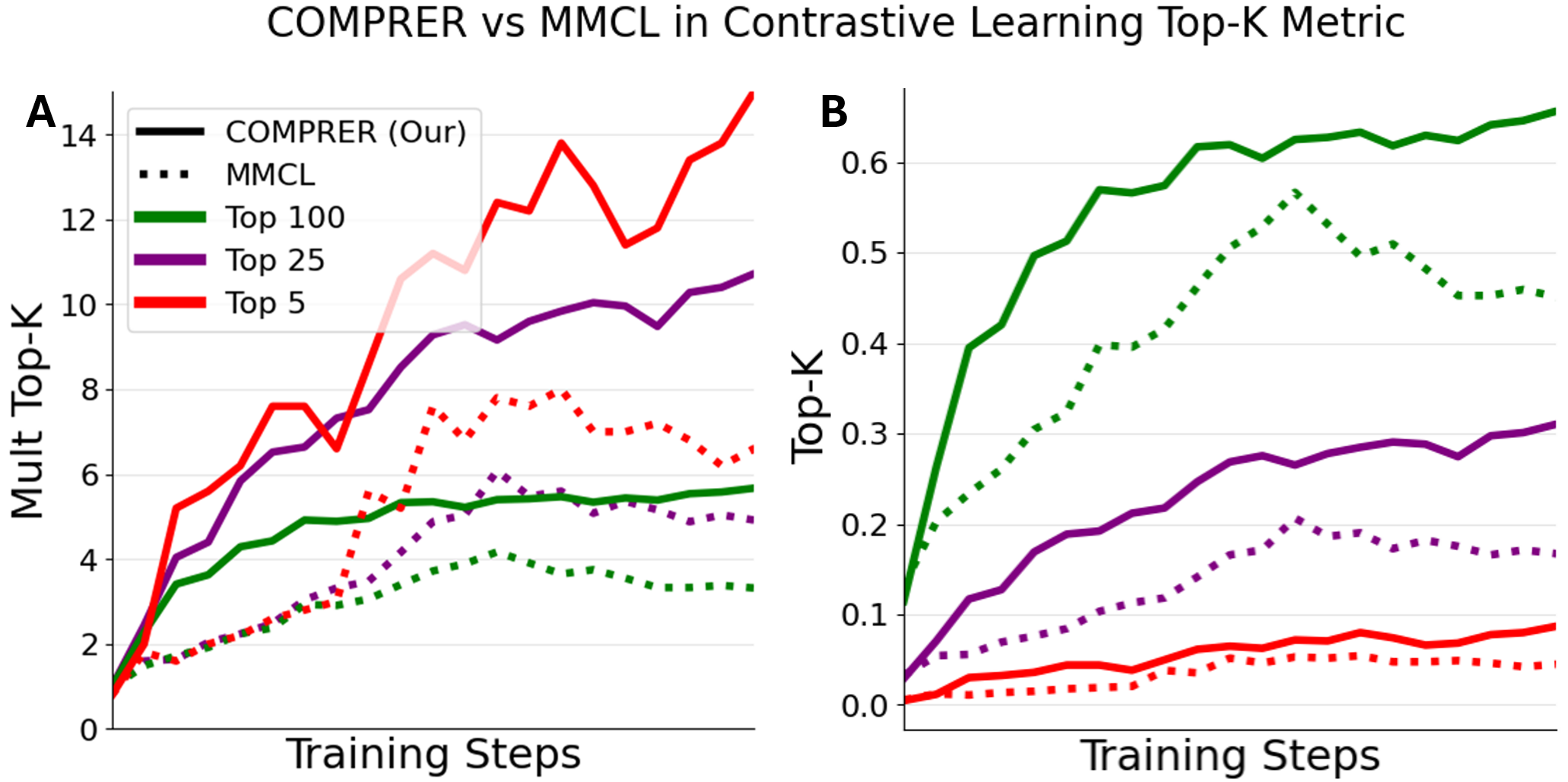} 
\caption{Comparative Analysis of Top-K Contrastive Metric on COMPRER and MMCL. We see in green, purple, and red the Top-100, Top-25, Top-5 contrastive metric respectively. We see in a dotted line the MMCL model, which trained solely on the multi-modal contrastive loss (trained using only  $\mathcal{L}_{\text{contr\_rc}}$  ), and in the straight line COMPRER which trained on multiple objectives, including the multi-modal contrastive loss. In \textbf{3.A} We see the multiplicative tok-K metric, and in \textbf{3.B} we see the top-K metric. We see that COMPRER consistently outperforms MMCL.}
\label{fig.3}
\end{center}
\end{figure}
In figure 3.b we see that all start off with random performance (mult top-K score of 1 for all K) , and rise as the optimization starts. In COMPRER, the observed Top-100 accuracy reached 0.65, which is higher than the baseline set by MMCL of 0.56. For the more stringent Top-25 accuracy, COMPRER obtained an accuracy of 0.35, while the MMCL achieved a lower score of 0.2. \textbf{This difference not only underpins our model's superior matching capability but also implies the potential benefits of multi-objective training in augmenting the feature space for more nuanced discriminatory powers.}

\subsubsection{Image Reconstruction Capability}
% Figure 4 presents a qualitative comparative analysis between the original test set images and their reconstructions from latent embeddings.
Both fundus image and carotid image reconstructions, while losing some fine details, remain structurally similar to the original general structure, an encouraging sign for the model's comprehension of microvascular features. While select losses in high-frequency details were observed—likely attributable to the inherent information compression within the network—the structural integrity was maintained.

% \begin{figure}[!ht]
% \begin{center}
% \includegraphics[scale=0.35]{recon.png} 
% \caption{Qualitative Assessment of Image Reconstruction Using The Image Decoder. This figure presents a side-by-side comparison of original and reconstructed medical images from a test set. The top row, labeled 'Original Images', displays a fundus image on the left and a carotid ultrasound image on the right. The bottom row, labeled 'Reconstructed Images', shows the image decoder's output. }
% \label{fig.4}
% \end{center}
% \end{figure}

\begin{figure}[!ht]
\begin{center}
\includegraphics[scale=0.32]{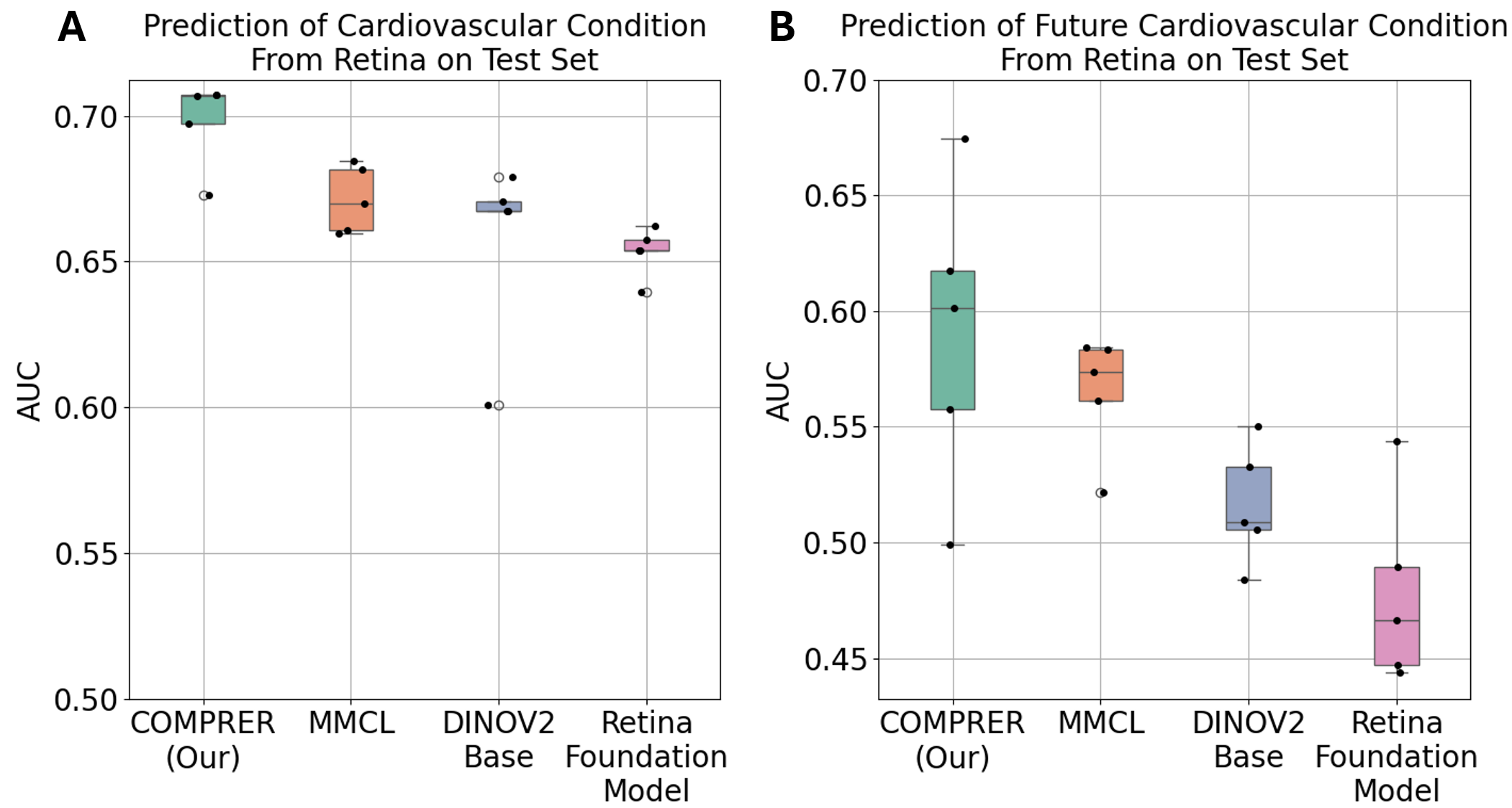} 
\caption{Comparative Analysis of Model Performance in Predicting Cardiovascular Conditions from Fundus Images. \textbf{4.A} illustrates the AUC results for various models fully finetuned to predicting current cardiovascular condition based on fundus images taken at the participants' first visit. The models compared include, our proposed COMPRER, MMCL, DINOV2-Base, and a Retina Foundation Model. The boxplots represent the distribution of AUC scores achieved on the test set after selection of top 5 hyperpameters for each model based on validation performance. \textbf{4.B} depicts the AUC results for the same set of models, but focuses on predicting the future onset of cardiovascular conditions. This analysis only includes participants who were healthy at baseline, with the goal of determining if the models can predict the development of cardiovascular conditions at a follow-up visit. While the AUC scores on their own are not high, COMPRER consistently outperforms the competition. It's also worth observing that MMCL has also some gains with respect to DINOV2 and the retina foundation model, which goes to show that even just applying multi-modal contrastive learning on its own can increase downstream performance.}
\label{fig.4}
\end{center}
\end{figure}

\subsection{Predictive Performance on Cardiovascular Conditions in the HPP Cohort}
While internal validation schemes during the pretraining phase provide essential insights into the immediate learning dynamics of our COMPRER model, the true test of its effectiveness lies in its clinical application. Thus, our goal is to demonstrate that our model pretraining not only captures intricate data patterns but also translates into significant improvements in real-world clinical diagnostics. To this end, we focused our attention on fine-tuning COMPRER to predict cardiovascular health conditions. Figure 4 shows the model’s capacities in both a diagnostic and prognostic context, providing valuable insights into cardiovascular health. The performance metrics presented in Figure 4 were constructed using the models, which was fitted with a 1-layer MLP regression head. To find appropriate hyperparameters, we employed a systematic hyperparameter search on the validation set, from which the top 5 models of each type were identified. These leading models were then assayed on an independent test set, yielding a distribution of results, which denotes the robustness and consistency of performance across model instances. The MMCL model is an ablation model, representing a version of our architecture and data trained with only multimodal contrastive loss, to provided a baseline to quantify the value added by the multi-objective learning. The Retina Foundation Model embodies a high-parameter (300M parameters, which is $ 3.5\times$ larger than all other competitor models) alternative, leveraging a considerably larger latent space (1024, which $\approx 1.3$ times larger than all other competitors) and trained on an extensive dataset of 1.6M fundus images (which is $75\times$ larger than our fundus dataset). Despite these advantages of the retina foundation model, COMPRER shows superior performance. We also observed that on the prognosis task (figure 4.b) almost all model runs except COMPRER's are random. This is interesting because COMPRER is the only model that had in its pretraining any signal of future events - based on the temporal contrastive learning objective.

\begin{figure}[ht]
\begin{center}
\includegraphics[scale=0.4]{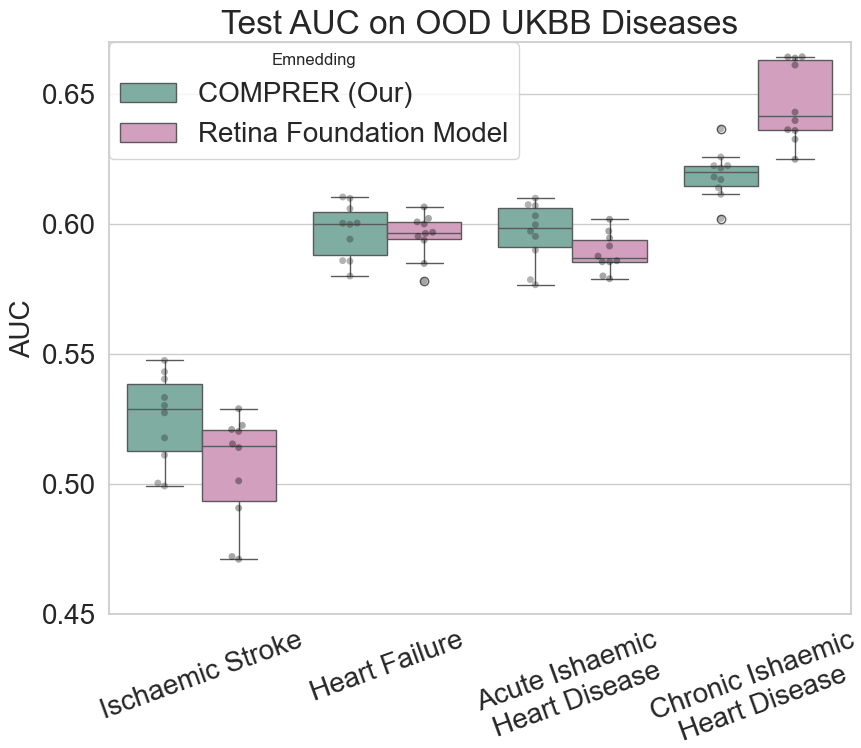} 
\caption{Comparative Analysis of COMPRER and Retina Foundation Model Performance in Predicting Diseases from Fundus Image Representations on an External Dataset (UKBB). The prediction is done from frozen image embeddings using logistic regression. We see that COMPRER outperforms the Retina Foundation model in the prediction of ischaemic stroke, evidenced by a higher AUC value. Conversely, the Retina Foundation model exhibits superior performance in predicting chronic ischaemic heart disease, although COMPRER also shows a test AUC above 0.6. For the conditions of stroke, heart failure, and acute ischaemic heart disease, the COMPRER method slightly outperforms the Retina Foundation model, though the margins are not significant. Its also notable that the the retina foundation model was trained using $75\times$ more fundus images, and with a $3.5\times$ larger model.}
\label{fig.5}
\end{center}
\end{figure}

% \begin{wrapfigure}{r}{0.5\columnwidth} % "r" for right side
%     \centering
%     \includegraphics[width=0.48\columnwidth]{actual_icml_24/ukbb_rev_rev4.png}
%     \caption{Comparative Analysis of COMPRER and Retina Foundation Model Performance in Predicting Diseases from Fundus Image Representations on an External Dataset (UKBB). The prediction is done from frozen image embeddings using logistic regression. We see that COMPRER outperforms the Retina Foundation model in the prediction of ischaemic stroke, evidenced by a higher AUC value. Conversely, the Retina Foundation model exhibits superior performance in predicting chronic ischaemic heart disease, although COMPRER also shows a test AUC above 0.6. For the conditions of stroke, heart failure, and acute ischaemic heart disease, the COMPRER method slightly outperforms the Retina Foundation model, though the margins are not significant. Its also notable that the retina foundation model was trained using $75\times$ more fundus images, and with a $3.5\times$ larger model.}
%     \label{fig.5}
% \end{wrapfigure}

\subsection{Performance on an OOD External Cohort}
Validating the predictive power of a model on an out of distribution (OOD) dataset is often considered the gold standard for demonstrating the real-world applicability and robustness of a predictive framework. In this section, we evaluate COMPRER's performance on the external dataset - the UK Biobank (UKBB), an extensive, well-characterized external cohort that has been at the forefront of large-scale biomedical research \citep{sudlow2015ukbiobank}. After data filtering and cleaning, it comprises of 44K participants with fundus images. The Retina Foundation model has previously showcased its traction on this dataset, establishing a performance benchmark for the field. Figure 5 illustrates the comparison between COMPRER and the Retina Foundation model in predicting various cardiovascular and related diseases from fundus image representations. Our approach demonstrates competitive, if not superior, predictive performance across numerous conditions. In the realm of ischaemic stroke prediction, COMPRER conspicuously outperforms the Retina Foundation model, denoting a higher AUC value. These results affirm the appropriateness of COMPRER's multi-modal, multi-objective pretraining paradigm, reinforcing its utility in extracting salient features pertinent to disease states from medical imagery. Moreover, they underscore our method's efficiency; by achieving these competitive performance metrics, COMPRER evidences that well-conceived model architectures coupled with sophisticated pretraining strategies can level the playing field against models with ostensibly more advantageous training conditions (using $75\times$ the data, and $3.5\times$ the parameters).

\section{Discussion}

We presented COMPRER, a novel pretraining method targeted at extracting medical features, specifically aimed at the enhancement of downstream tasks such as disease diagnosis and prognosis. Our multi-modal, multi-objective pretraining approach bridges the gap between advancing machine learning architectures and their practical applications in clinical settings, exhibiting superior predictive abilities compared to specialized models. In the realm of image representation, COMPRER has demonstrated the ability to distill meaningful insights from a large-scale, longitudinally derived dataset, diminishing the reliance on arduously curated labels. The empirical advancements evidenced by robust internal validation and insightful applications in disease prediction advocate for COMPRER's adoption. Notably, the model provides competitive performance, transcending the need for vast data volumes and extensive computational resources, highlighting a quality-centric approach in medical data analytics. We have shown that the integration of multiple objectives stretches the training duration but enriches the model's feature representations. This comprehensive approach, though intensive, equips the model with a deepened understanding and enhanced performance that may surpass models optimized for single objectives alone. The superior performance in multimodal contrastive matching demonstrates this phenomenon, suggesting that a more holistic, integrated training regimen can indeed yield models with robust generalization capacities across various tasks.

In addition, we have shown that COMPRER is capable of achieving higher test AUC scores than models that have been trained with an order of magnitude more data, and with larger models. We have shown this both in our HPP cohort, as well as, in an external cohort - the UKBB, which shows our model can generalize out of distribution, across diverse populations from different continents. Interestingly, within the OOD UKBB dataset, a notable divergence in the test AUC was observed for ischaemic stroke prediction between COMPRER, and the Retina Foundation model. This finding is particularly intriguing given that ischaemic stroke is a condition diagnosable through carotid ultrasound \citep{zhang2014carotid}.  It is worth noting that COMPRER was trained using both fundus images and carotid ultrasound, suggesting that this multi-modality approach may have played a role in the increase of COMPRER's predictive accuracy compared to the Retina Foundation model. It is also noteworthy that in the internal cohort validation section (5.2), where we showed future prognosis performance, the majority of model runs, with the exception of COMPRER, exhibit random test AUC scores. This phenomenon might be interesting, as it underscores the distinctive attribute of COMPRER's pretraining methodology. In contrast to the other models, COMPRER benefits from the inclusion of a temporal information during its pretraining, notably - the temporal, visit-based, contrastive learning objective. This objective provided the model with insights into future events within the longitudinal data. We hypothesize that this unique aspect of COMPRER's training regimen is contributing to its superior performance, even surpassing models like MMCL that have encountered the same data but lack exposure to the temporal, visit-based contrastive loss. This finding underscores the potential advantages of incorporating temporal information in pretraining, shedding light on the nuances of disease prognosis prediction and highlighting the efficacy of our approach.

However, our work has several caveats, calling attention to the simultaneous challenges and potential trajectories for improvement. Dataset scope and representativeness remain pivotal for model generalization. While the HPP dataset underpins our current findings, incorporating datasets with broader demographic diversity is essential for enhancing model robustness and ensuring its translational relevance across patient populations. A more diverse dataset would help mitigate bias and uphold the model's diagnostic integrity, especially when encountered with dataset shifts in real-world scenarios. The generalizability of COMPRER across different diseases and imaging modalities is another frontier to be explored. Expanding the disease spectrum and experimenting with a variety of modalities are key to consolidating the framework's applicability in diverse medical contexts. Moreover, the limited computational resources constrained our model’s training to 40k steps, hinting at the potential for further refinement. Investments in computational infrastructure and collaborative efforts could uncover latent performance enhancements and insights into the optimization dynamics of our model. Interpretability and explainability are indispensable for clinician and patient acceptance. Despite the strides made with our transformer-based model, elucidating the AI's decision-making processes remains crucial. Validation in clinical environments could unravel the efficacy and adaptability of our model and solidify its role within clinical workflows. It is important to note that we conducted our research on only two modalities, however, we do see how this method could easily extend beyond only two modalities. In the HPP dataset, we have access to a rich variety of over 20 distinct data modalities, ranging from visual information and time series data to textual records and tabular measurements. 

We recognize the untapped potential of leveraging multiple modalities within our pretraining scheme. Future iterations of the methods we described here can harness this diverse data landscape by incorporating additional losses for multi-modal contrastive learning, introducing multi-visit losses that span across these various modalities, and exploring other innovative techniques. This multi-modal approach holds promise in further enriching the model's understanding of complex medical data, potentially leading to even more robust generalization across a wide array of clinical tasks.

\section{Availability of Code and Model Weights}
In the interest of transparency and facilitating future research, we plan to release the code and model weights associated with our study upon publication.

\nocite{*}
\section{Bibliographical References}\label{sec:reference}

\bibliographystyle{unsrtnat}
\bibliography{ref}

\end{document}